\documentclass{article}

\PassOptionsToPackage{round}{natbib}

     \usepackage[final]{neurips_2025}

\usepackage[utf8]{inputenc} 
\usepackage[T1]{fontenc}    
\usepackage{hyperref}       
\usepackage[dvipsnames]{xcolor}         
\definecolor{midnightblue}{rgb}{0.098,0.098,0.439}
\hypersetup{
    colorlinks=true,
    linkcolor=midnightblue,
    citecolor=midnightblue,
    urlcolor=midnightblue
}
\usepackage{url}            
\usepackage{booktabs}       
\usepackage{amsfonts}       
\usepackage{nicefrac}       
\usepackage{microtype}      
\usepackage{xcolor}         
\usepackage{todonotes}

\usepackage{comment}
\usepackage{hyperref}
\usepackage{url}
\usepackage{comment}
\usepackage{listings}

\usepackage{graphicx}
\usepackage{amssymb,amsmath}
\usepackage[capitalize]{cleveref}

\title{Investigating Representation Universality:\\ Case Study on Genealogical Representations}

\author{%
  David D. Baek \\
  MIT\\
  \texttt{dbaek@mit.edu} \\
  \And
  Yuxiao Li \\
  Independent\\
  \texttt{}\\
  \And 
  Max Tegmark \\
  MIT \\
  \texttt{tegmark@mit.edu}\\
}

\def\ie{{\frenchspacing\it i.e.}}

\def\A{\textbf{A}}
\def\B{\textbf{B}}
\def\E{\textbf{E}}

\def\p{\textbf{p}}

\def\H{\textbf{H}}

\def\K{\textbf{K}}
\def\U{\textbf{U}}

\def\R{\textbf{R}}

 \def\E{{\bf E}}
 \def\R{\mathbb{R}}

\def\spose#1{\hbox to 0pt{#1\hss}}
\def\simlt{\mathrel{\spose{\lower 3pt\hbox{$\mathchar"218$}}
     \raise 2.0pt\hbox{$\mathchar"13C$}}}
\def\simgt{\mathrel{\spose{\lower 3pt\hbox{$\mathchar"218$}}
     \raise 2.0pt\hbox{$\mathchar"13E$}}}
\def\simpropto{\mathrel{\spose{\lower 3pt\hbox{$\mathchar"218$}}
     \raise 2.0pt\hbox{$\propto$}}}

\begin{document}

\maketitle

\begin{abstract}
Motivated by interpretability and reliability, we investigate whether large language models (LLMs) deploy universal geometric structures to encode discrete, graph-structured knowledge. To this end, we present two complementary experimental evidence that might support universality of graph representations. First, on an in-context genealogy Q\&A task, we train a cone probe to isolate a ``tree-like'' subspace in residual stream activations and use activation patching to verify its causal effect in answering related questions. We validate our findings across five different models. Second, we conduct model stitching experiments across models of diverse architectures and parameter counts (OPT, Pythia, Mistral, and LLaMA, 410 million to 8 billion parameters), quantifying representational alignment via relative degradation in the next-token prediction loss. Generally, we conclude that the lack of ground truth representations of graphs makes it challenging to study how LLMs represent them. Ultimately, improving our understanding of LLM representations could facilitate the development of more interpretable, robust, and controllable AI systems.
\end{abstract}

\section{Introduction}




\def\mt#1{{\bf [Dowon: #1]}}

Large Language Models (LLMs), despite being primarily trained for next-token predictions, have shown surprisingly robust reasoning capabilities~\citep{bubeck2023sparks,claude_3_tech_report,gemini_tech_report}. 
However, despite recent progress, we still lack a clear understanding of how these models internally encode different kinds of knowledge. Improving such understanding could enable valuable progress relevant to transparency,  interpretability, and safety; For example,
(a) discovering and correcting inaccuracies to improve model reliability \citep{zhang2024self}, (b) discovering and correcting biases \citep{chen2024large}, (c) revealing and removing dangerous knowledge \citep{zhang2024adversarial}, and (d) detecting deceptive behavior where models deliberately output information inconsistent with its knowledge \citep{marks2023geometry}.

Prior works have identified geometric structures of specific kinds of knowledge in LLMs and shown that these structures recur across many different models -- evidence of representation universality. For example, \cite{gurnee2023language} identified a linear subspace that captures spatio-temporal coordinates; \cite{engels2024not} discovered a circular manifold of calendar days and months' representations; and \cite{kantamneni2025language} demonstrated a helical subspace of number representations. However, the question of how LLMs represent discrete, relational structures -- such as nodes and edges in a knowledge graph -- remains largely unexplored.
In this paper, we ask:
\begin{center}
    
\emph{Do LLMs exhibit representation universality when encoding graph‐structured knowledge?}
\end{center}

To investigate representation universality for discrete, graph-structured knowledge, we present two complementary experimental evidence:

1. \textbf{Tree-structured subspace of Genealogy representations}:
When representing \emph{descendant-of} relationship, we identify that the optimal representation is a tree-like embedding that could be identified via cone probe.
On an in‐context genealogy Q\&A task, we use a cone probe to isolate a tree‐like subspace within the residual stream activations. We then use activation patching to verify the causality of this subspace. We validate our findings across five different models.

2. \textbf{Cross-model alignment via Model Stitching}: Since we lack a ground-truth representation for arbitrary graphs, we adopt a black-box model stitching approach to compare representations across different LLMs. We splice the early layers of one model onto the late layers of another via trainable linear adapter. Our experiments cover a diverse set of models -- from OPT and Pythia to Mistral and LLaMA -- ranging in size from 410 million to 8 billion parameters. By measuring the increase in next-token prediction loss relative to each model’s baseline, we quantify representation alignment between different models.

Together, these experiments provide supporting evidence that LLMs may employ universal geometric structures to represent graphs.
The rest of this paper is organized as follows:  In Section \ref{prob_form}, we formally describe the problem setting and introduce our hypothesis for the optimal representation of genealogical relationships. In \cref{sec:llm-geometry}, we investigate whether LLM representations exhibit geometric structure similar to the optimal representation we propose.  \cref{sec:llm-stitching} presents additional evidence for representational universality via LLM stitching experiments. We relate our approach to prior work in \cref{sec:related-work}, and conclude our paper in \cref{sec:conclusion}.

\begin{figure}
    \centering
    \includegraphics[width=.49\linewidth]{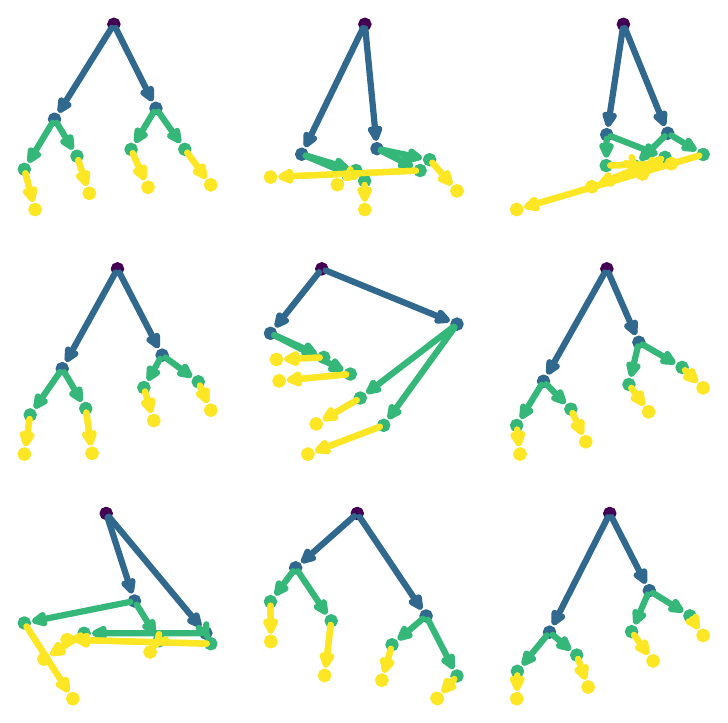}
    \includegraphics[width=.49\linewidth]{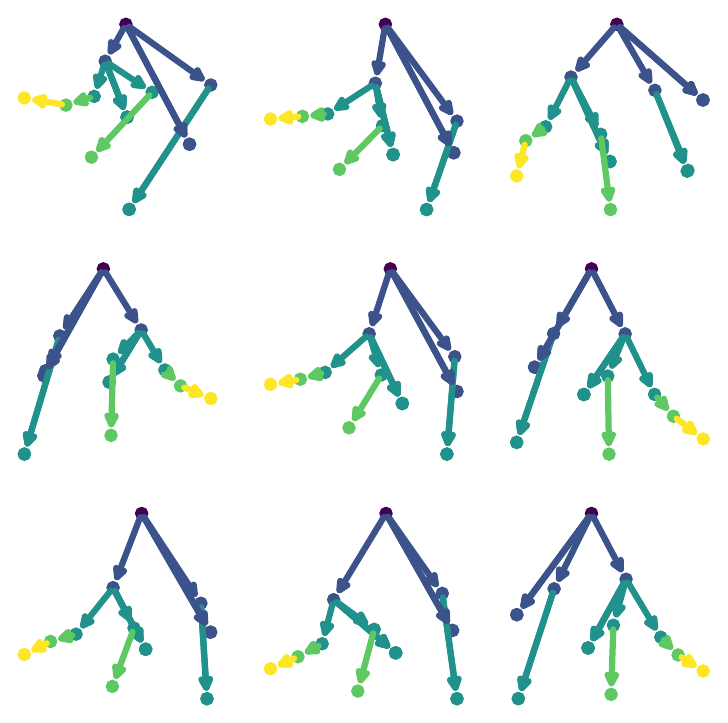}
    \caption{Visualization of the top two principal components of an MLP trained to learn the \emph{descendant-of} relationship across nine different random seeds -- for models trained on either (left) a fully balanced binary tree or a (right) randomly generated general tree consisting of 15 nodes. For clarity, we add arrows connecting direct parent–child pairs. Each plot is rotated so that the root node appears at the top of the panel. Across different seeds and tree structures, the learned representations consistently exhibit a geometric pattern that resembles a tree in discrete mathematics -- a structure we define as \emph{cone embedding} in the main text. Note that the models do not separate two sibling leaf nodes under the same parent. This is because all embeddings are initialized to zero, and the model receives no gradient signals to separate two sibling leaf nodes -- they are equivalent nodes when it comes to determining the \emph{descendant-of} relationship. }
    \label{fig:mlp-probes}
\end{figure}

\section{Setup}
\label{prob_form}
Consider a general knowledge graph (KG) consisting of $m$ binary relations $R^{(1)}, R^{(2)}, \cdots R^{(m)}$ between $n$ objects (nodes) $x_1,...,x_n$. Our task is to understand the representation that enables link prediction, the task of predicting the probability $p_{ijk}$ that $R^{(i)}(x_j,x_k)=1$. While most KG-learning algorithms in the literature embed both objects and relations \citep{cao2024knowledge}, we instead embed only objects ($x_i\mapsto\E_i\equiv \E(x_i)\in\R^d$) and train a link predictor network  
$\p(\E_j,\E_k)$, which takes two embedding vectors $\E_j,\E_k$ as an input, and outputs an $m$-dimensional vector $\p$ that represents link probability. This is to emulate the behavior of modern large language models, where only objects are embedded and relations are implicitly defined via weights. Our ultimate goal is to improve our understanding of representations that enable knowledge graph learning tasks in LLMs. 

As a specific instance of this problem, consider a problem of learning \emph{descendant-of} relationship in a tree.
We claim that the optimal representation of this problem is a \emph{cone} embedding, where $j$ is a direct descendant of $i$ iff $\E_j$ lies within a fixed cone emanating from $\E_i$.\footnote{By 
 \emph{optimal}, we mean a representation that satisfies all the special properties of the relation, such as symmetricity and transitivity. We discuss optimal representations in more detail in \cref{app:opt-rep}.} We show that cone embedding is the optimal representation for this problem in \cref{app:opt-rep}. Accordingly, we could define a score function which measures the probability that $j$ is a direct descendant of $i$:
\begin{equation}
    \p(\E_{i},\E_{j}) = \sigma (\E_{i,1}-\E_{j,1}) \sigma(\E_{i,0}-\E_{j,0}),
\end{equation}

where $\sigma$ is a sigmoid function and $\E_{i,n}$ denotes the $n-$th component of embedding $\E_i$. Since this score function is differentiable, we could also train a probe that measures how close a given embedding is to the cone embedding, which we refer to as the \emph{cone probe}.

To test this hypothesis, we train a multi-layer perceptron (MLP) with a single hidden layer of width 50 to learn the \emph{descendant-of} relationship on a tree consisting of 15 nodes. We do not use a test split, as our primary goal is to analyze the geometric structure of representations rather than to evaluate their generalization performance.
The model embeds each object into a two-dimensional space, concatenates the resulting vectors, and passes them through the MLP to predict the probability that node $j$ is a direct descendant of node $i$. We use the AdamW optimizer \citep{loshchilov2017decoupled} with a learning rate of $10^{-3}$, and train for up to $10^4$ epochs, while applying early stopping if the loss does not improve for 30 consecutive epochs. We perform experiments on both a fully balanced binary tree and a randomly generated general tree.

\cref{fig:mlp-probes} visualizes the top two principal components of the learned embeddings across nine different random seeds, both for the balanced tree and the general tree, with arrows added from each parent to its child for clarity. We observe that the learned representations indeed form \emph{cone embeddings} -- a geometric structure that closely resembles tree-like hierarchies in discrete mathematics. Another notable observation is that the model organizes the representations into a meaningful geometric structure, even though it could, in principle, simply memorize the training data and has no explicit incentive to learn a structured embedding. We hypothesize that this emergent structure is driven by the model's dimensionality constraint -- specifically, the requirement to encode all relevant information within a two-dimensional space. This limitation effectively forces the model to arrange the objects into a coherent tree-like layout.

In the following section, we investigate whether genealogical representations in LLMs exhibit similar geometric structure. We will use the cone probe to identify relevant subspaces and perform causal interventions on them.

\begin{figure}[t]
    \centering
    \includegraphics[width=\linewidth]{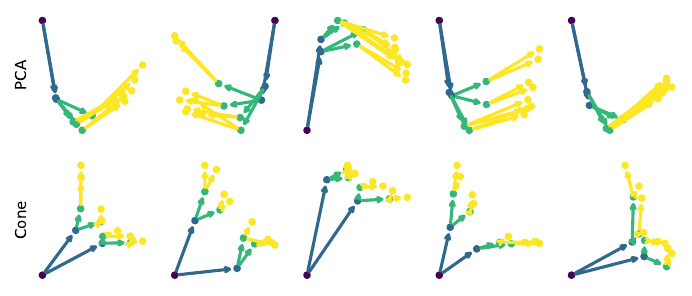}
    \includegraphics[width=\linewidth]{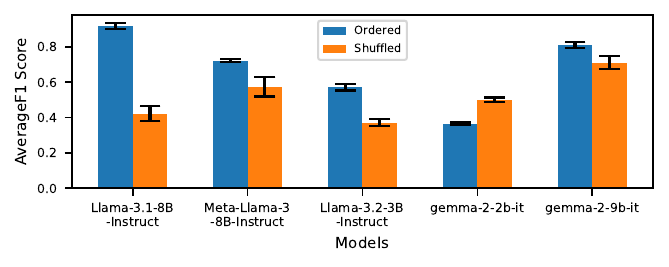}
    \caption{\textbf{Top}: Visualization of in-context genealogy-tree representations from LLaMA-3.1-8B-Instruct across five different random name assignments on a full binary tree of 15 nodes. We show the projection onto the first two principal components, and the Projection onto the cone-probe subspace. Nodes and edges are colored by their depth in the tree. We added arrows connecting direct parent-to-child links for visualization. \textbf{Bottom}: Average F1 score on question-answering tasks about \emph{descendant-of} relationships, averaged over five different name assignments on a tree. These results suggest that the model may struggle with compositional generalization if the relevant facts are not provided in order.}
    \label{fig:llm-pca}
\end{figure}
\section{Genealogy Representations in LLMs}
\label{sec:llm-geometry}

In the previous section,  we observed that small models often encode genealogical relationships in a tree-like structure. This raises an interesting question: would LLMs represent genealogies in a similar way? To investigate, we design an in-context genealogy task as follows. We generate a full binary tree with 15 nodes and assign each node a name drawn at random (without replacement) from the 200 most popular male and female names from the birth year 2000, using the \texttt{pybabynames} package in Python \citep{pybabynames2024}. In the prompt, we first describe the family tree by listing all the children of every person on each line. We then ask questions of the form ``Is \emph{X} a direct descendant of \emph{Y}?'' to the LLM. We show an example of the full prompt in \cref{app:full-prompt}. We evaluate our results over five different models, which are listed in \cref{app:list-models}.

\cref{fig:llm-pca} shows the average F1 score on question-answering tasks about \emph{descendant-of} relationships, averaged over five different name assignments on a tree.  First, we found that the models are only able to answer these questions well when the lines, each of which lists the children of a specific person, are ordered based on the person's depth in the tree. When the orders are randomly shuffled, the model's performance significantly deteriorated. This is in accordance with the well-known reversal curse \citep{berglund2023reversal}, where LLMs trained on ``A is B'' fail to learn ``B is A''. Hence, if we present ``C is a child of D'' first, and then ``B is a child of C,'' the model may not be able to identify the reverse compositional relationship that ``B is a child of D.'' Hence, we focus on studying the representations of the family tree when the graph descriptions are ordered based on people's depth in the tree.

To identify tree-like subspaces, we train a cone probe on the residual stream activations at the target token. To prevent overfitting, we first reduce each activation vector to 10 dimensions via PCA and then fit the cone probe in this lower dimensional space. We train a cone probe with AdamW optimizer with a learning rate $10^{-3}$ for 3000 epochs, while keeping the model that achieves the best F1 score on the original dataset. \cref{fig:llm-pca} visualizes the resulting 2D embeddings from PCA projection and cone projection across five different family trees (i.e. names are assigned to each node at random). We find that the PCA representations tends to be more degenerate (nodes at the same depth cluster tightly), whereas the cone probe yields a clearer, discrete ``branching'' structure that mirrors the underlying tree topology.

\begin{figure}[t]
    \centering
    \includegraphics[width=\linewidth]{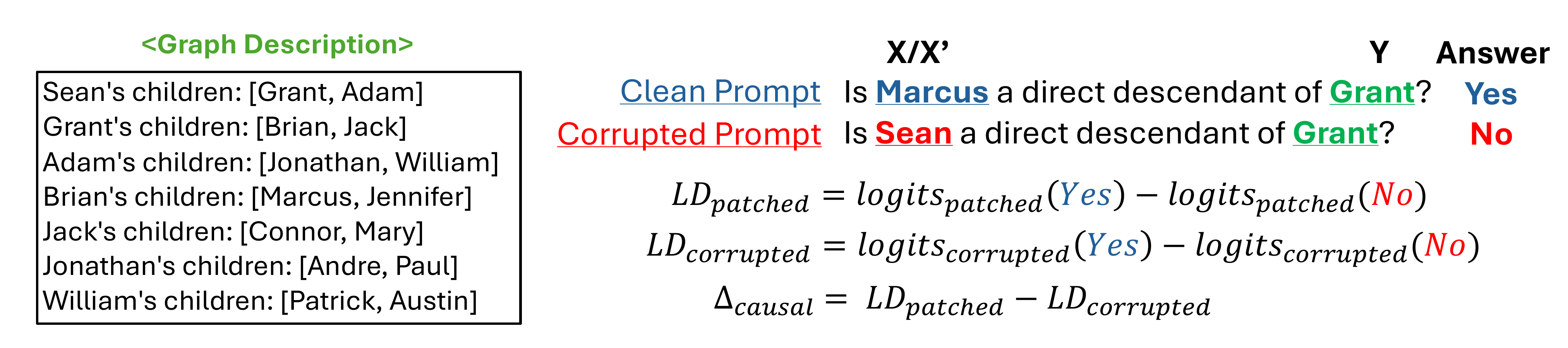}
    \includegraphics[width=\linewidth]{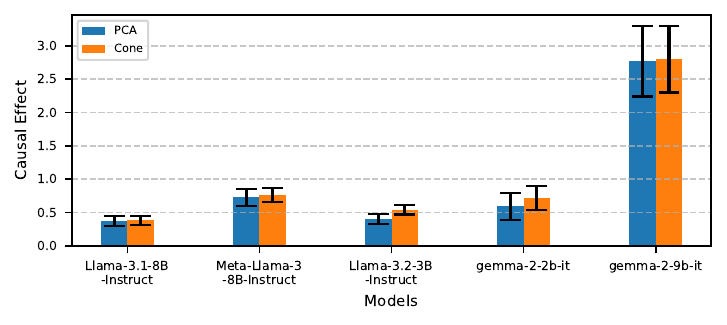}
    \caption{\textbf{Top}: Illustration of our intervention methodology. \textbf{Bottom}: Intervention results across five models. The histogram shows the causal effect of patching two subspaces of the residual stream activations at one-third model depth: (a) the subspace spanned by the top two principal components and (b) the cone subspace. }
    \label{fig:patching-result}
\end{figure}

To verify the causal role of these subspaces, we conduct an intervention experiment. For each family tree, we sample 100 prompt pairs from five trees with different name assignments -- a ``clean'' prompt (\emph{X}, \emph{Y}) and a corresponding ``corrupted'' prompt (\emph{X'}, \emph{Y}) -- constructed so their correct answers are opposite. We balance the set so that half of the clean prompts yield a positive (Yes) answer and the other half yield a negative (No) answer. For each pair, we run the model on the corrupt prompt, while patching its residual stream activations at layer $l$ with those recorded from the clean run. We then quantify the causal effect of patching by comparing logit differences:

\begin{equation}
    \Delta_{\rm causal} = LD_{\rm patched} - LD_{\rm corrupted},
\end{equation}

where $LD_x$ is the logit difference between the correct and incorrect tokens in run $x$. We compare three activation patching scenarios: (a) Patching the full layer, (b) Patching the top two principal components, and (c) Patching the cone subspace. More precisely, suppose \(B \in \mathbb{R}^{d\times k}\) spans the subspace of interest, and define the orthogonal projection matrix
\[\mathbf{P} \;=\; B\,(B^\top B)^{-1}B^\top.\]
For any representation \(\mathbf{x}\in\mathbb{R}^d\), the patched representation is given by

\begin{equation}
    \mathbf{x}_{\rm patched} = \mathbf{x}_{\rm corrupted} - \mathbf{P}\mathbf{x}_{\rm corrupted} + \mathbf{P}\mathbf{x}_{\rm clean}
\end{equation}

\begin{figure}[t]
    \centering
    \includegraphics[width=\linewidth]{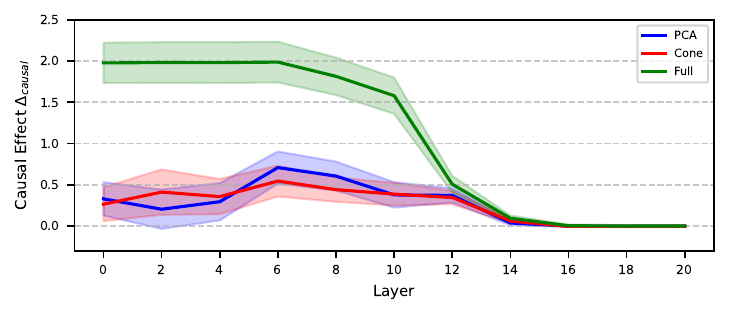}
    \caption{Intervention results for Llama-3.1-8B-Instruct across different layers. The plot shows the causal effect of patching two subspaces of the residual stream activations: (a) the subspace spanned by the top two principal components and (b) the cone subspace. Standard errors are indicated as a shaded region. Full represents patching the full activation at a specific layer.}
    \label{fig:patching-layer}
\end{figure}

Intervention results are shown in \cref{fig:patching-result} and \cref{fig:patching-layer}. We find that representations in early to mid layers exhibit a stronger causal effect than those in later layers. Moreover, patching the cone‐probe subspace alone produces a logit shift that is comparable to or larger than patching the top two principal components. Although full-layer patching yields an even larger effect -- implying additional causally relevant directions beyond the cone subspace -- our findings confirm that the cone subspace reliably emerges when models are asked to answer a question about a tree of relatively small size, and is at least as causally relevant as the subspace spanned by the top two principal components.


\textbf{Limitations:} First, we focus solely on the internal geometry and universality of LLM representations -- without examining how these subspaces are actually leveraged by the model for answering questions, more well known as circuit analysis \citep{tigges2024llm}. Second, our experiments use a relatively small binary tree consisting of 15 nodes on which models achieve near-perfect accuracy in answering related questions. Consequently, it remains unclear whether similar causal, tree-like subspaces emerge in larger or more complex genealogies, or if future, more capable models will encode genealogies in a similar manner. For instance, for a particular task of answering \emph{descendant-of} questions, the ratio between positive and negative samples approaches zero as the tree size approaches infinity. Therefore, the model might just learn to say No for all questions while still getting accuracy larger than 99\%. Hence, we would need a model that is good at what is known as the \emph{needle-in-a-haystack} problem \citep{liu2023lost}.

\section{LLM Stitching Experiments}
\label{sec:llm-stitching}

\subsection{Model Stitching}

Model Stitching \citep{lenc2015understanding,bansal2021revisiting} is a method for probing the representation similarity between two different models by constructing a hybrid model that \emph{stitches} the bottom layers of one model to the top layers of another model via trainable adapter layer. By measuring the performance drop of the stitched model relative to the original model, one could infer the degree of representation alignment between two different models. In this section, we apply this method to LLMs to study representation alignment between different LLMs.

Formally, the process of stitching two LLMs could be described as follows: Consider two LLMs
\begin{equation}
    \A=\U^A \left(\prod_{i=0}^{n-1}\H_i\right)\E^A, \; \B=\U^B \left(\prod_{i=0}^{m-1}\K_i\right)\E^B,
\end{equation}
 where $\H_i, \K_i$ are decoder layers, $\E$ is the embedding layer, and $\U$ is the unembedding layer. The stitched model is given by
\begin{equation}
    \B \circ \A = \U^B \left(\prod_{i=(m-l+1)}^{m-1}\K_i\right) S(\Lambda)\left(\prod_{i=0}^{k-1}\H_i\right) \E^A,
\end{equation}
where we stitched the first $k$ layers of $\A$ and the last $l$ layers of $\B$. We then train a linear stitching layer $S(\Lambda)$ to minimize the next-token prediction cross-entropy loss:

\begin{equation}
    \mathcal{L}(\Lambda) = \sum {\rm log} \left[\mathbb{P}(v_i | v_{i-1}\cdots v_0, \Lambda)\right].
\end{equation}

For stitching models with different tokenizers, $v_i$ is the first token of the string $v_iv_{i+1}v_{i+2}\cdots$, tokenized by $\B$'s tokenizer.
For our experiments, we stitch models from the OPT family (1.3B, 2.7B, 6.7B), Pythia family (410M, 1.4B, 2.8B), Mistral-7B-Instruct, and LLaMA-3.1-8B-Instruct. These models were chosen to cover a wide spectrum of model families and parameter scales. We trained the stitching layer for 10,000 steps with a linearly decaying learning rate starting at $10^{-3}$ with 100 warmup steps, and a weight decay of $10^{-4}$. We used open-source models available in Huggingface, and used Huggingface datasets \emph{monology/pile-uncopyrighted} and \emph{monology/pile-test-val} for training and evaluating test loss. Each sample is truncated to 2048 tokens, and we report average test loss over 2000 randomly selected test samples.

\subsection{Results}

\begin{figure}
    \centering
    \includegraphics[width=\linewidth]{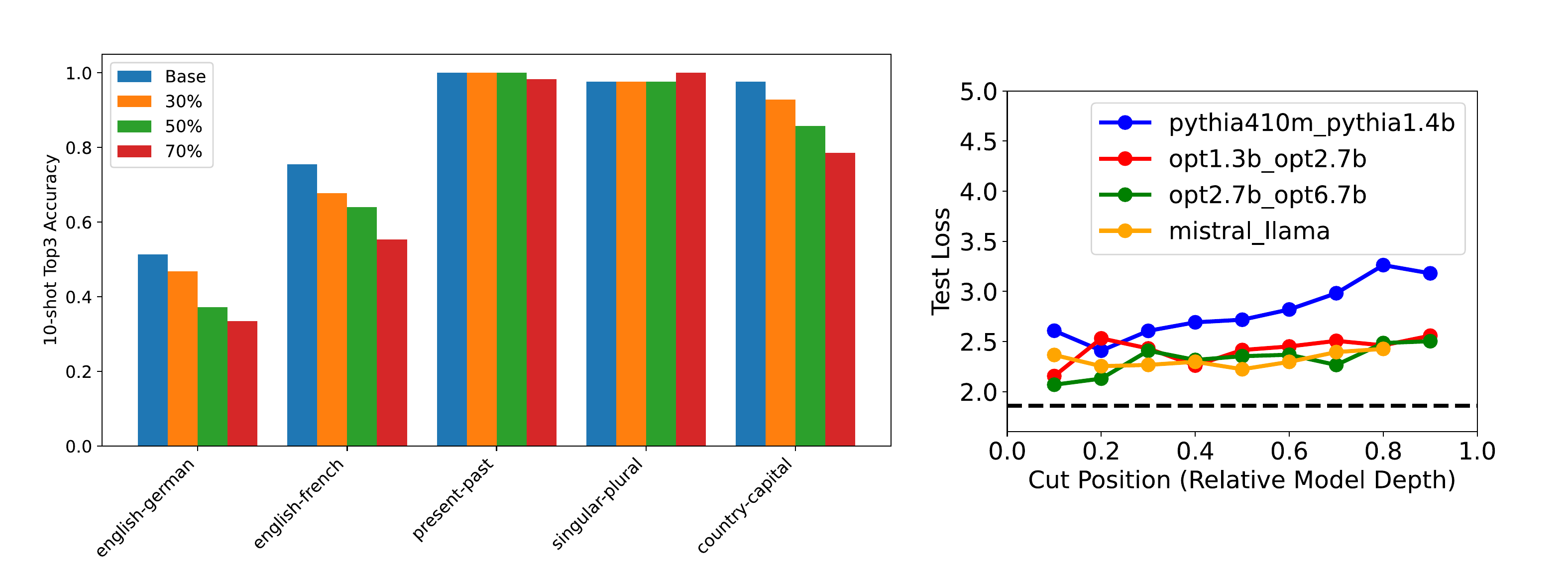}
    \caption{\textbf{Left}: In-context learning accuracy for models stitched between OPT-2.7B and OPT-6.7B. Base indicates OPT-6.7B, and $x$\% indicate the embedding layer and first $x$\% of the OPT-6.7B replaced by those of OPT-2.7B. \textbf{Right}: Test loss as a function of stitched position between two different models. The two models are cut at the same relative depth within each model. The black dashed line on the right figure indicates the average test loss of original models.}
    \label{fig:llm-stitching}
\end{figure}

\cref{fig:llm-stitching} presents the results of our LLM stitching experiments. Overall, we observe that representations from different models align more closely in early to mid layers than in later layers. Correspondingly, in-context learning performance declines as the stitching point moves to later relative depths.

\begin{figure}
    \centering    
    \includegraphics[width=.49\linewidth]{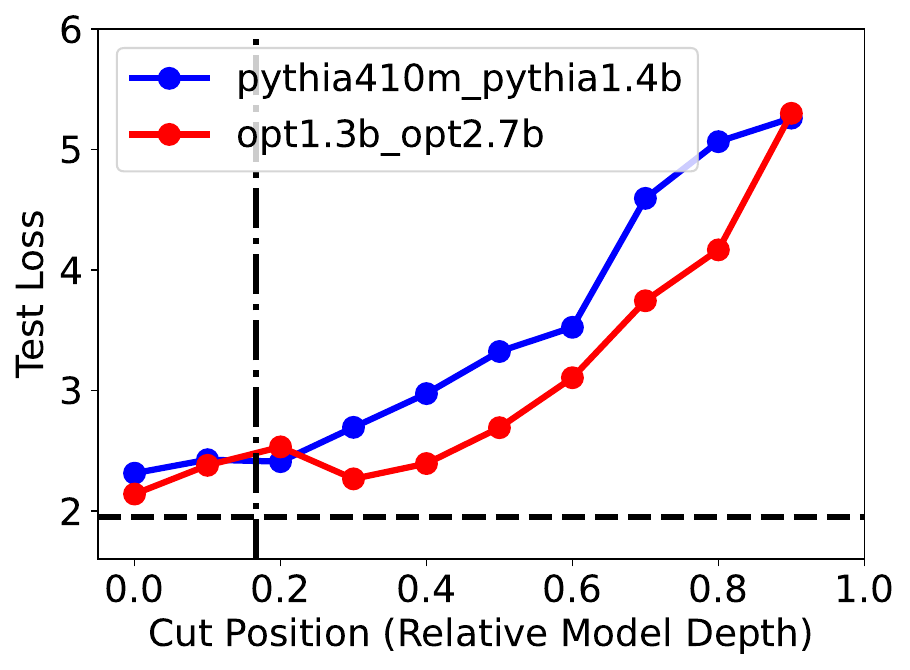}
    \includegraphics[width=.49\linewidth]{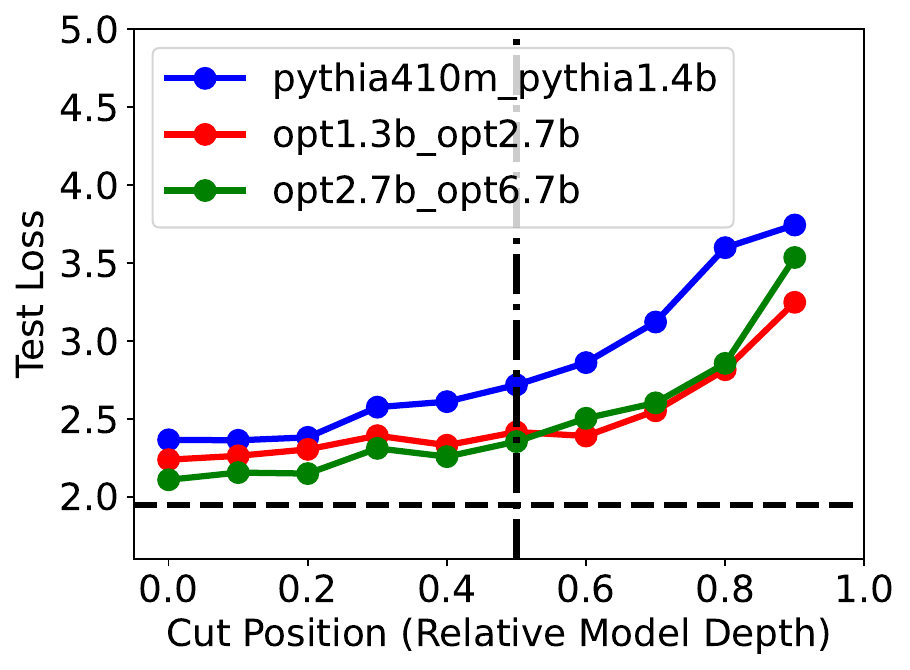}
    \caption{Test loss as a function of the stitch point $x$. We stitch the first $x$\% of layers from the second model onto the first model at two fixed depths: one-sixth of its total depth (left) and one-half of its total depth (right). The vertical dashed line marks the relative depth where the first model is cut (one-sixth and one-half, respectively). The horizontal dashed line indicates the original models’ average test loss.}
    \label{fig:llm-stitching-fixed}
\end{figure}

\begin{figure}
    \centering
    \includegraphics[width=.49\linewidth]{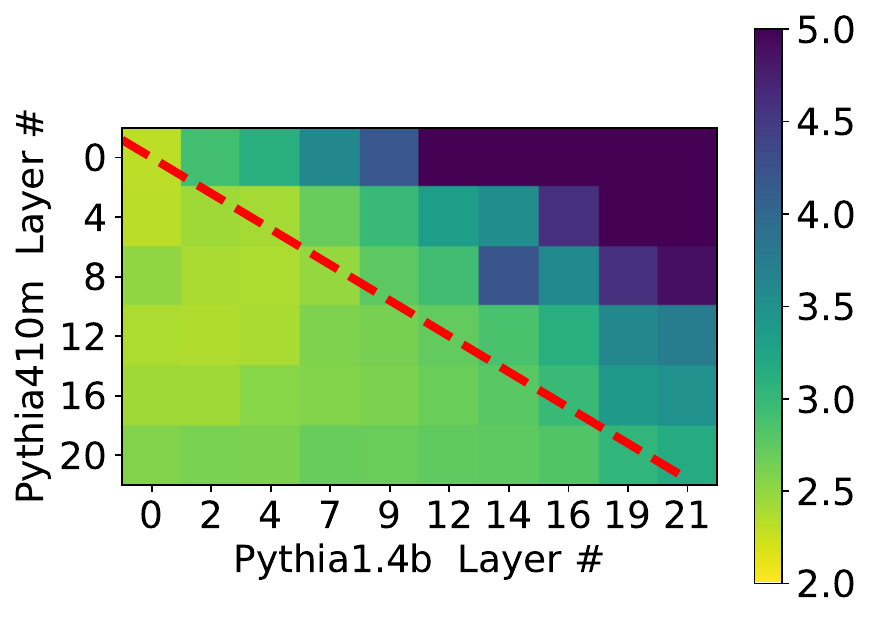}
    \includegraphics[width=.49\linewidth]{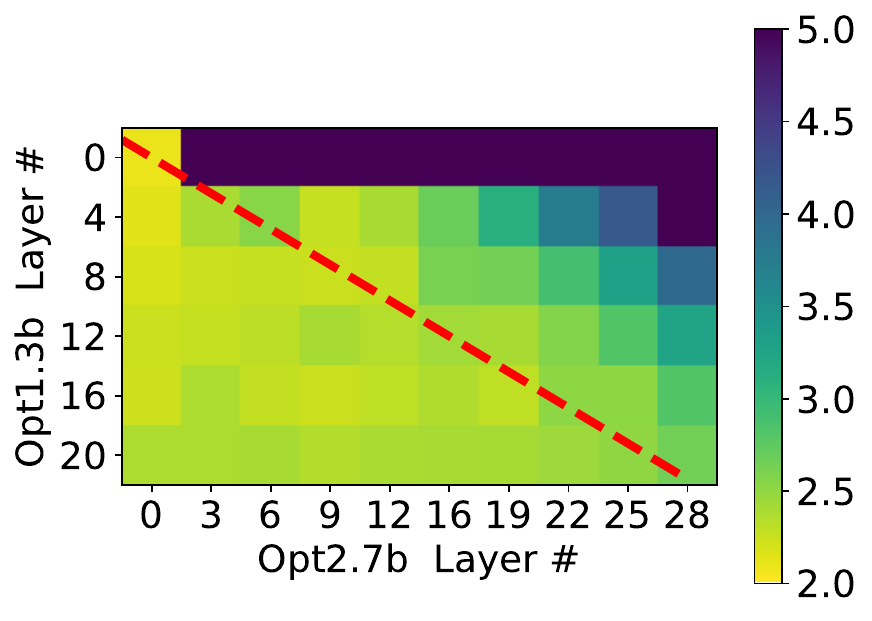}
    \caption{Test loss for different stitching configurations. Each point $(i,j)$ indicates the loss of the model obtained by taking the first $j$ layers of the $y$-axis model and the last $(L-j)$ layers of the $x$-axis model, where $L$ is the total number of layers of the $x$-axis model. The red diagonal line marks the cases where both models are joined at the same relative depth.}
    \label{fig:llm-stitching-table}
\end{figure}

We also evaluated stitching various layers of one model onto a fixed layer of another (\cref{fig:llm-stitching-fixed} and \cref{fig:llm-stitching-table}). Test loss remains relatively low when connecting the embedding layer of one model to downstream layers of another, suggesting substantial representational transformations in the first few layers as token-level embeddings are converted into higher-level semantic concepts. While mid-layer representations between models are often compatible, stitching them into later layers yields higher loss -- likely because those layers prioritize next-token prediction over forming semantic concepts. Interestingly, we can stitch a mid-layer of one model onto an early layer of another (e.g., layers 0–15 of Pythia-410M to layers 2–23 of Pythia-1.4B), implying that mid-layer activations still retain sufficient token-level information which could be ``reset'' to token-level representations.

These results corroborate the \emph{Stages of Inferenece} hypothesis of \cite{lad2024remarkable}, which argues that LLMs process inputs through discrete phases -- first constructing semantic representations in early-to-mid layers, then shifting to next-token prediction in later layers. Consequently, representations at equivalent relative depths across different models exhibit strong alignment, the concept known as representation universality \citep{huh2024platonic}.

\textbf{Limitations:} Despite its utility in quantifying representational alignment, our experiment has a few limitations. First, it assumes that representational alignment can be completely captured through a simple linear mapping; therefore, more complex or nonlinear representations, such as circular features in days of the month \citep{engels2024not} or helical features in numbers \citep{kantamneni2025language}, may be classified as not equivalent. In order to circumvent this problem, one could consider adding a quadratic correction term to the adapter layer. Moreover, our experiments span only a limited set of LLM architectures and scales; therefore, it may not generalize to other models, such as multimodal models, that are not studied in this paper.

\section{Related Works}
\label{sec:related-work}

In light of the recent development of LLMs' capabilities, understanding the inner workings of Large Language Models have become increasingly important to ensure the safety and robustness of AI systems \citep{tegmark2023provably,dalrymple2024towards}.

\textbf{Mechanistic Interpretability} Neural Networks have demonstrated a surprising ability to generalize \citep{liu2021towards,ye2021towards}. Recently, there have been increasingly more efforts on trying to reverse engineer and interpret neural networks' internal operations \citep{zhang2021survey,bereska2024mechanistic,baek2024geneft}. Such methods include using structural probes and interventions at the level of entire representations \citep{hewitt2019structural,pimentel2020information}, and studying neuron activations at the individual neuron level \citep{dalvi2019one,mu2020compositional}. Our work is part of this broader effort in mechanistic interpretability; We aim to understand how large language models represent different types of knowledge.

\textbf{Knowledge Representations in Language Models} Early word‐embedding models, including Word2Vec and GloVe, were found to encode semantic relationships as linear directions in their vector spaces ~\citep{drozd2016word, pennington2014glove, ma2015using}. More recently, several studies showed that LLMs are capable of forming conceptual representations in spatial, temporal, and color domains \citep{gurnee2023language,abdou2021can,li2021implicit}. Some studies focused primarily on examining the linearity of LLMs' feature representations \citep{gurnee2023language,hernandez2023linearity}. Several works found multi-dimensional representations of inputs such as lattices~\citep{michaud2024opening} and circles~\citep{liu2022towards,engels2024not}, 
one-dimensional representations of high-level concepts and quantities in large language models~\citep{gurnee2023language, marks2023geometry, heinzerling2024monotonic,park2024geometry}.

In particular, \citet{park2024geometry} studied representations of word hierarchies. We examine representations of genealogical trees -- another form of hierarchical data but are fundamentally different from word hierarchies because individuals in different generations do not possess inherent semantic relationships. Our findings suggest the existence of more multi-dimensional features, warranting further investigation. Our work is closely related to \citet{park2024iclr}, who study representations developed during in-context learning on a graph-tracing task. However, their analysis focuses on lattice and ring structures, which are inherently one-dimensional. In contrast, we aim to study in-context learning representations arising from data with more complex, hierarchical structures.

Our work is also closely related to traditional knowledge graph embedding models such as TransE \citep{wang2014knowledge}, ComplexE \citep{trouillon2016complex}, and TransR \citep{lin2015learning}, which embed both entities and relations into a shared latent space and optimize a scoring function for link prediction. In contrast, our approach embeds only entities (objects), most closely mirroring how LLMs represent and process information.


\textbf{Representation Alignment and Model Stitching} There are active discussions in the literature about strengths and weaknesses of different representation alignment measures \citep{huh2024platonic,bansal2021revisiting,sucholutsky2023getting}. Several works have considered stitching to obtain better-performing models, such as stitching vision and language models for image and video captioning task \citep{li2019visual,iashin2020multi,shi2023learning}, and stitching BERT and GPT for improved performance in look ahead section identification task \citep{jiang2024look}. Some works have considered stitching toy transformers to understand the impact of activation functions on model's performance \citep{brown2023understanding}. Our work considers stitching LLMs to examine the hints of representation universality across different models.

\section{Conclusion}
\label{sec:conclusion}
We studied whether LLMs deploy universal geometric structures to encode graph-structured knowledge. We presented two complementary experimental evidence that supports universality of graph representations of LLMs. First, on an in-context genealogy Q\&A task, we trained a cone probe to isolate a ``tree-like'' subspace in residual stream activations and utilized activation patching to verify its causal effect in answering related questions. Second, we conducted model stitching experiments across diverse architectures and parameter counts, and quantified representational alignment via relative degradation on next-token prediction loss. Generally, we conclude that the lack of ground truth representations of graphs makes it challenging to study how LLMs represent them. Ultimately, improving our understanding of LLM representations could facilitate the development of more interpretable, robust, and controllable AI systems.

\textbf{Future Works:} One could systematically investigate the optimal representations of more complex genealogical relationships -- such as cousins, aunts, and uncles -- and analyze whether LLMs encode these relations in a similar geometric manner. It would also be interesting to explore whether there exists a critical graph size beyond which such optimal representations begin to emerge. Our current study is limited to relatively small graphs, since  model performance on genealogical question-answering tasks degrades significantly with increasing graph size. To address this, one could fine-tune existing LLMs or employ larger, more capable models to better understand the emergence of structured representations in larger graphs.

Another promising direction is to examine how LLMs estimate their uncertainty when reasoning over graph-structured data. We observe that LLMs rarely express full confidence in their answers to descendant-of questions, even for relatively small trees. Applying mechanistic interpretability techniques to study how uncertainty is represented could provide valuable insights into how LLMs process genealogical relationships in context.

\bibliography{neurips_2025}
\bibliographystyle{plainnat}

\newpage
\appendix


\section{Optimal Representation in Knowledge Graph Learning}
\label{app:opt-rep}

We define  \emph{optimal} representation as those that satisfiy all the special properties of the relation. Such properties include
\begin{itemize}
    \item Symmetricity: $\forall x_1, x_2:R(x_1,x_2)\implies R(x_1,x_2)$
    \item Reflexivity: $\forall x_1: R(x_1,x_1)=1$
    \item Transitivity: $\forall i,j,k:R(x_i,x_j)\land R(x_j,x_k)\implies R(x_i,x_k)$
    \item Meta-transitivity: $\forall i,j,k:R^{(1)}(x_i,x_j)\land R^{(1)}(x_j,x_k)\implies R^{(2)}(x_i,x_k)$
\end{itemize}

As an example, we prove that \emph{cone} embedding in the main text is an optimal representation of the \emph{descendant-of} relationship.

Proof. Our predictor function for cone probe is given by $\p(E_i,E_j)=H(E_{i0}-E_{j0})H(E_{i1}-E_{j1})$ where $H$ is the heaviside step function ($H(x)=1$ if $x>0$, vanishing otherwise). We show that $\p$ satisfies transitivity, \ie if $i$ is a descendant of $j$, and $j$ is a descendant of $k$, then $i$ is a descendant of $k$:

Suppose $\p(E_i,E_j)=\p(E_j,E_k)=1$.  By definition of the cone probe,
\[
\p(E_i,E_j)=1
\iff
E_{i0}>E_{j0}
\;\wedge\;
E_{i1}>E_{j1},
\quad
\p(E_j,E_k)=1
\iff
E_{j0}>E_{k0}
\;\wedge\;
E_{j1}>E_{k1}.
\]
Chaining these inequalities gives
\[
E_{i0}>E_{k0}
\quad\text{and}\quad
E_{i1}>E_{k1},
\]
and hence $\p(E_i,E_k)=H(E_{i0}-E_{k0})\,H(E_{i1}-E_{k1})=1$.\hfill $\square$

\section{Full Prompt Example}
\label{app:full-prompt}
\begin{lstlisting}
Below is an instruction that describes a task, paired with an input that provides further context. Write a response that appropriately completes the request.

### Instruction:
Answer a question about the family tree relationship based on the given data. If it's a yes/no question, answer with only one word: 'Yes' or 'No.' If it's a 'who' question, answer with the person's name(s).

### Input:
Family Tree:
Emily's children: [Scott, Jordan]
Scott's children: [Marco, William]
Jordan's children: [Charles, Hunter]
Marco's children: [Luke, Jose]
William's children: [Jessica, Crystal]
Charles's children: [Alan, Joseph]
Hunter's children: [Laura, Grace]

Question: Is Grace a direct descendant of Laura?

### Response:
\end{lstlisting}

\newpage
\section{List of Models}
\label{app:list-models}

\begin{table}[h]
    \centering
    \begin{tabular}{l|c}
        \toprule[0.4mm]
        \textbf{Model Name} & \textbf{Citation} \\
        \midrule[0.4mm]
        meta-llama/llama-3.1-8b-instruct   & \cite{touvron2024llama3} \\
        meta-llama/Meta-Llama-3-8B-Instruct  & \cite{touvron2024llama3} \\
        meta-llama/Llama-3.2-3B-Instruct  & \cite{touvron2024llama3} \\
        google/gemma-2-2b-it  & \cite{gemma2024} \\
        google/gemma-2-9b-it  & \cite{gemma2024} \\
        \bottomrule[0.4mm]
    \end{tabular}
    \vspace{2mm}
    \caption{List of Models used in our experiments.}
    \label{tab:ai-models-Elo}
\end{table}

\end{document}